\pdfoutput=1

\documentclass[11pt]{article}

\usepackage[]{acl}

\usepackage{times}
\usepackage{latexsym}

\usepackage[T1]{fontenc}

\usepackage[utf8]{inputenc}

\usepackage{microtype}

\usepackage{inconsolata}

\usepackage{amsmath}
\usepackage{graphicx}
\usepackage{float}
\usepackage{subcaption}
\usepackage{amsfonts}
\usepackage{multicol}
\usepackage[bottom]{footmisc}
\usepackage{diagbox}
\usepackage{bm}
\usepackage{bbm}

\usepackage{multirow, booktabs}
\usepackage{caption}
\usepackage{color, colortbl}
\definecolor{Gray}{gray}{0.9}

\usepackage{pifont}
\newcommand{\cmark}{\ding{51}}%
\newcommand{\xmark}{\ding{55}}%

\newcommand{\argmax}[1]{\underset{#1}{\text{argmax}} \;}

\title{LLM Comparative Assessment: Zero-shot NLG Evaluation through Pairwise Comparisons using Large Language Models}

\author{Adian Liusie, Potsawee Manakul, Mark J. F. Gales \\
  ALTA Institute, Department of Engineering, University of Cambridge \\
  \texttt{al826@cam.ac.uk, pm574@cam.ac.uk, mjfg@eng.cam.ac.uk}}

\begin{document}
\pagenumbering{arabic} 
\maketitle
\begin{abstract}
Current developments in large language models (LLMs) have enabled impressive zero-shot capabilities across various natural language tasks. An interesting application of these systems is in the automated assessment of natural language generation (NLG), a highly challenging area with great practical benefit. In this paper, we explore two options for exploiting the emergent abilities of LLMs for zero-shot NLG assessment: absolute score prediction, and comparative assessment which uses relative comparisons between pairs of candidates. Though comparative assessment has not been extensively studied in NLG assessment, we note that humans often find it more intuitive to compare two options rather than scoring each one independently. This work examines comparative assessment from multiple perspectives: performance compared to absolute grading; positional biases in the prompt; and efficient ranking in terms of the number of comparisons. We illustrate that LLM comparative assessment is a simple, general and effective approach for NLG assessment. For moderate-sized open-source LLMs, such as FlanT5 and Llama2-chat, comparative assessment is superior to prompt scoring, and in many cases can achieve performance competitive with state-of-the-art methods. Additionally, we demonstrate that LLMs often exhibit strong positional biases when making pairwise comparisons, and we propose debiasing methods that can further improve performance.

\end{abstract}

\section{Introduction}
With the current rapid advances in generative AI, pre-trained models are increasingly utilized in a range of NLP tasks, necessitating reliable evaluations of these models. Human evaluation, where annotators critically assess the quality of the outputs of natural language generation (NLG) systems, has been the gold standard approach \cite{lita-etal-2005-blanc, belz-reiter-2006-comparing, lai-tetreault-2018-discourse, fabbri2021summeval}. However, human evaluation has its drawbacks, and is notably labor-intensive, time-consuming, and costly. As such, automating the evaluation process and assessing NLG systems without human intervention is highly desirable. 

\begin{figure}[t]
    \centering
    \includegraphics[width=\columnwidth]{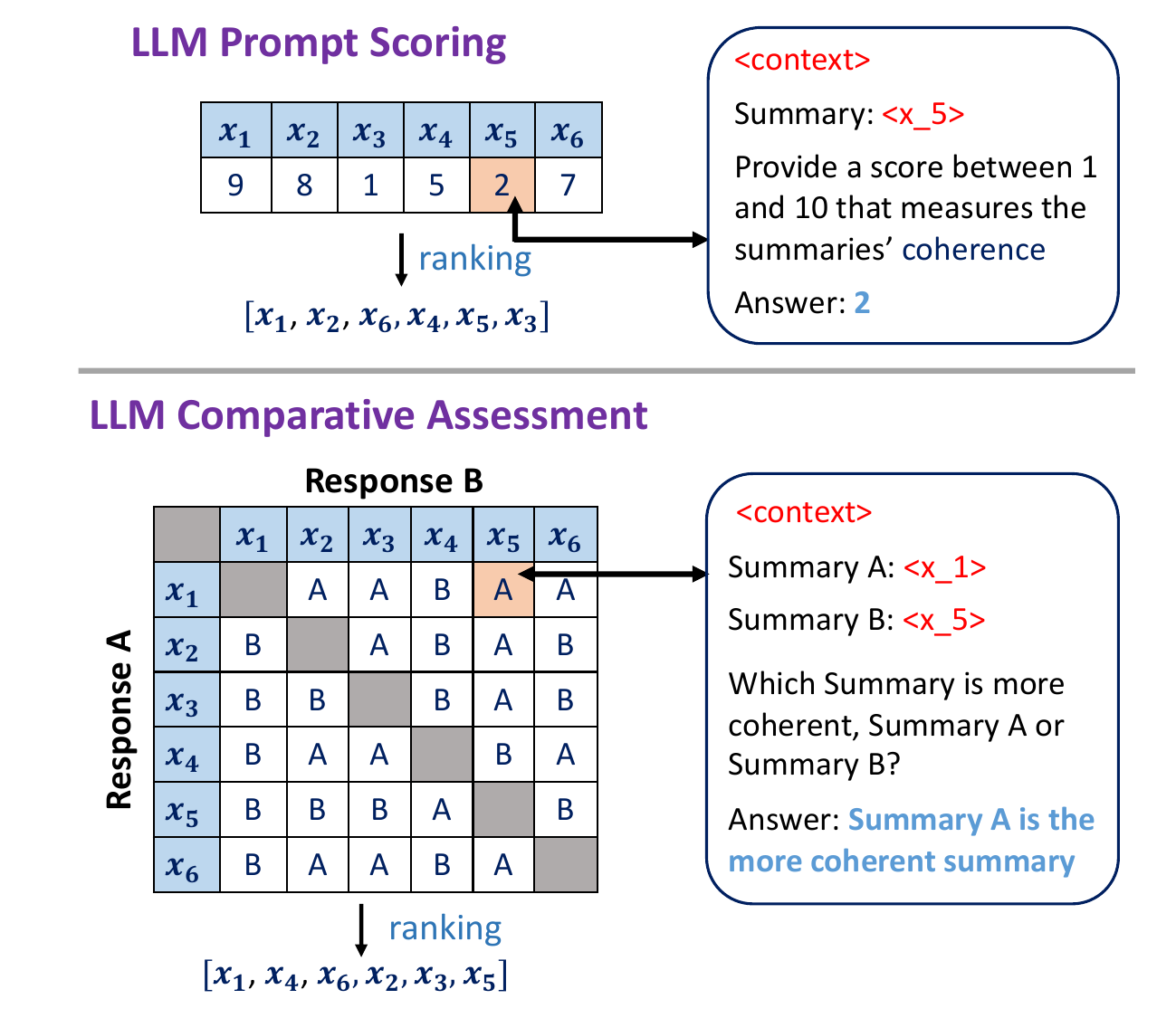}
    \caption{Prompt Scoring v.s. Comparative Assessment. Comparative Assessment prompts an LLM to compare candidates in a pairwise manner, and the comparisons are subsequently converted into scores or ranks.}
    \label{fig:main-diagram}
\end{figure}

Though there has been considerable progress in automatic evaluation methods, many proposed approaches have certain restrictions that limit their effectiveness. A large body of existing work use evaluation methods designed for particular tasks and attributes \cite{mehri-eskenazi-2020-unsupervised, rei-etal-2020-comet, manakul2023selfcheckgpt}, for example, measuring the consistency of summaries \cite{wang-etal-2020-asking, manakul2023mqag}. Though effective within their domain, these approaches are not extensible to different NLG aspects and cannot be used by practitioners wishing to evaluate systems on inputs or properties that are less common. 

The recent development in the emergent abilities of LLMs \cite{wei2022emergent} has enabled LLMs to achieve impressive zero-shot performance for a slew of language tasks. This has led to general prompt-based assessment approaches, such as prompt-scoring where an LLM is probed to score outputs on a particular aspect \cite{wang2023chatgpt, kocmi2023large}. These approaches are often only effective with massive LLMs with 175B+ parameters, which may limit the applicability of the approach, especially when access is limited to API access.  

With the insight that for humans, it is often easier to select which of two options is better than it is to score options independently, we question whether pairwise comparisons may be more effective at leveraging the impressive emergent ability of LLMs. In this work, we consider LLM comparative assessment, where an LLM is prompted to compare pairs of NLG candidates and predict which one is better. We demonstrate empirically that comparative assessment performs much better than prompt-scoring for FlanT5 and Llama style models, and enables moderate-sized open-source LLMs to achieve near (or above) state-of-the-art performance across a range of NLG language tasks, for a diverse set of attributes. Our approach is general and can be applied to a diverse range of tasks and textual attributes, is simple and requires minimal prompt engineering. Further, we demonstrate that pairwise LLM comparisons often exhibit strong positional biases, where the ordering of candidates impacts the decisions. We introduce a simple debiasing method and empirically illustrate that debiasing can provide further performance improvements, especially when large biases are present. 

Our contributions are 1) We are the first work that comprehensively analyzes pairwise comparative assessment for NLG evaluation; 2) We demonstrate that comparative assessment is far more effective than prompt-scoring for moderately-sized LLMs, and yields performance that is state-of-the-art for particular attributes; 3) We demonstrate that positional bias impacts comparative decisions, and introduce a method to debias LLMs which leads to performance boosts, especially when only a subset of comparisons are considered.  


\section{Background and Related Work}

\subsection{Reference-based Evaluation} 
In NLG evaluation, a standard approach is the comparison of annotator-provided gold-standard references with the generated response. Established heuristics, such as the N-gram overlap metrics ROUGE \cite{lin-2004-rouge} and METEOR \cite{banerjee-lavie-2005-meteor}, have extensively been applied for assessing summarization and machine translation respectively. Recently, the paradigm has evolved to incorporate embedding-based methods like BERTScore \cite{zhang2019bertscore}, which not only compares generated texts with references, but also factors in semantic considerations beyond word overlap.



\subsection{Tailored NLG Evaluation Approaches} 
Tailored approaches have been proposed for assessing specific properties of generated texts. For example, question-answering systems are used for summary consistency assessment \cite{wang-etal-2020-asking, scialom-etal-2021-questeval} to probe information consistency. For Dialogue quality assessment, the language model probability from a DiaoloGPT system is used as a proxy for response quality \cite{mehri-eskenazi-2020-usr}. A survey for NLG evaluation methods was conducted by \citet{celikyilmaz2020evaluation}. 


\subsection{Zero-shot LLM Evaluation}
Given the current capabilities of LLMs such as ChatGPT and GPT4, the zero-shot ability of these systems for a wide range of tasks, including NLG evaluation, has been investigated. Existing works have looked at using LLM to evaluate open-ended story generation and adversarial attacks \cite{chiang-lee-2023-large} and using ChatGPT to score the quality of texts along a certain axis \cite{wang2023chatgpt, kocmi2023large}, demonstrating that ChatGPT can be used in a zero-shot setting and achieve reasonable performance. 

\subsection{LLM Pairwise Comparisons}
Pairwise comparative judgement \cite{thurstone1927law} has been a popular approach of assessing candidates for exams, however where typically human assessors are used. Investigating the ability and application of pairwise comparisons via LLMs has been relatively underexplored, with concurrent work using pairwise rankings for information text retrieval \cite{qin2023large} and separately for assessing LLM-based chat assistants on open-ended questions where outputs are compared to that of a baseline system \cite{vicuna2023, zheng2023judging}. 

\section{Comparative Assessment}
\label{section:comparative_assessment}
\subsection{Notation}
In this work, we investigate using LLM comparative judgements for NLG assessment. Assume that there is a context $d$ (e.g., a text passage or dialogue) and a set of $N$ candidate responses, $x_{1:N}$. For a given attribute (e.g., coherence, consistency, fluency) the $N$ candidates have true underlying scores, $s_{1:N}$. As scores often only have relative meaning, in this work only the ranks of the candidates will be evaluated. The objective is therefore to accurately predict the true ranks, $r_{1:N}$, of the candidate scores.  In comparative assessment, one uses pairwise comparisons to determine which of the two input responses is better. Let $y_{ij} \in \{0,1\}$ represent the true outcome of whether $x_i$ is higher ranked than $x_j$, such that $y_{ij} = \mathbbm{1}(s_i > s_j)$. Here, an LLM is used to model the probability that response $i$ is better than response $j$,  $p_{ij}$,
\begin{align}
    p_{ij} = P(y_{ij} | x_i, x_j, d)
\end{align}
Which can alternatively be converted into hard decisions, $\hat{y}_{ij}$, by selecting the most likely outcome. 
\begin{align}
    \hat{y}_{ij} = \begin{cases}
	1, & \text{if $p_{ij} > 0.5$}\\
        0, & \text{otherwise}
    \end{cases}
\end{align}
Let $\mathcal{C}=\{c_k\}_{k=1...R}$ represent a set of comparisons, where $R$ is the total number of comparisons, and each comparison $c=(i, j)$ indicates the indices of the two considered candidate responses. For example,  the set of all possible comparisons, $\mathcal{C} =\{(i, j) \; | \; i, j \in [1...N], i\ne j \}$, could be used, or alternatively a smaller subset of comparisons. 

\subsection{Prompt Design}
\begin{figure}[H]
    \centering
    \includegraphics[width=0.7\columnwidth]{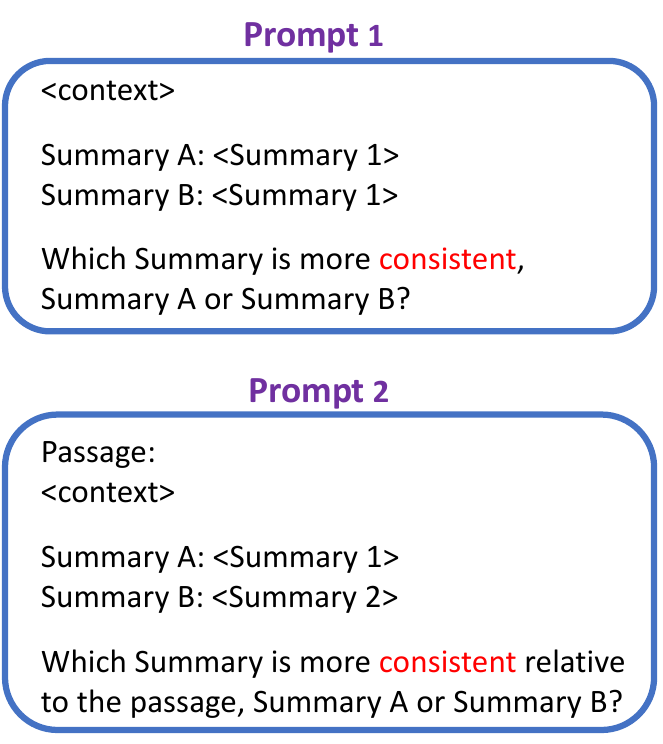}
    \caption{Comparative prompt template 1 and 2. When assessing different attributes, only the attribute is changed (e.g., consistent $\rightarrow$ engaging) and for response assessment, the word `summary' is replaced with `response'.}
    \label{fig:comparative_prompts}
\end{figure}

To leverage the emergent ability of LLMs, we use comparative prompts that probe a model to decide which of the two candidates is better. Let $T$ be a prompt template that converts candidate responses $x_i$ and $x_j$ as well as context $d$ into an output text, prompt $\mathcal{P} = T(x_i, x_j, d)$. This work aims to find a simple, general and robust assessment method, and as such extensive prompt engineering is not in the scope of this work (despite possible performance gains). We evaluate two simple and suitable prompts in our initial investigations. Our prompts for comparative assessment are shown in Figure \ref{fig:comparative_prompts}. 

\subsection{Comparative Decisions}
\label{section:comparative_decision}
A central aspect of LLM comparative assessment is the methodology of getting comparative decisions. In this section, we consider two approaches for leveraging LLMs for comparative assessment; First for when one has output token-level probabilities (Prompt-Based Classifier), and second for when only the output texts are available. 

\vspace{2mm}

\noindent\textbf{Prompt-Based Classifier}:
If one has access to the output probabilities, an efficient method to get probability estimates of the predictions is to leverage prompt-based classifiers. Let $P_{\theta}(w|x)$ represent an LLM's conditional language model distribution of the output sequence $w$ given the input text $x$. For prompt-based classifiers, the LM probabilities of specific label words ($w_k$) are used as a proxy for the class decisions~\cite{liusie2023mitigating}. For example in summarization assessment, given a prompt $\mathcal{P}$ ending in `... which summary is better', one can set $w_i$=`\textit{Summary A}' and $w_j$=`\textit{Summary B}' and define the probability that response i is better than response j as:
\begin{equation}
    p_{ij} = \frac{P_{\theta}(w_i|\mathcal{P})}{P_{\theta}(w_i|\mathcal{P}) + P_{\theta}(w_j|\mathcal{P})}
    \label{eq:argmax}
\end{equation}

\vspace{1mm}

\noindent\textbf{Text Generation}: 
Alternately, if only limited API access is available, one can sample responses from the conditional LM given the input prompt $\mathcal{P}$,
\begin{equation}
    \tilde{w}^{(k)} \sim P_{\theta}(w|\mathcal{P})
\end{equation}
Let $f(\tilde{w}) \in \{0,1\}$ be a function that maps the text response to the comparative decision. By generating $K$ samples from the LLM, one can estimate the comparative probability $p_{ij}$ by looking at the fraction of the samples that selects $x_i$ over $x_j$.
\begin{align}
    p_{ij} = \frac{1}{K} \sum_{k=1}^{K} f(\tilde{w}^{(k)})
\end{align}

\subsection{Comparisons to Ranks}
\label{sec:comp_select}
Although the full set of possible comparisons yields the most information for the rankings, this requires $R\!\!=\!\!N(N\!\!-\!\!1)$ comparisons, which can be computationally expensive. For computational efficiency, we can consider 3 different comparison selection strategies: random, no-repeat and symmetric. For \textbf{random}, comparisons are randomly selected from the set of all possible comparisons. For \textbf{no-repeat}, if $(x_i, x_j)$ is selected then $(x_j, x_i)$ will not be selected. For \textbf{symmetric}, if $(x_i, x_j)$ is selected, then $(x_j, x_i)$ will also be selected. 

Given a set of selected comparisons $\mathcal{C}$ and weights of a comparative assessment system $\bm{\theta}$, one can generate a predicted rank ordering $\hat{r}_{1:N}$ of the candidate responses. A simple but effective approach is to sort the candidates by \textbf{win-loss} ratio,
\begin{equation}
    \hat{s}_i = \frac{\text{\#wins of }x_i}{\text{\#comparisons involving } x_i}
\end{equation}
which can then be ordered to convert the scores into predicted ranks $\hat{r}_{1:N}$. 

\subsection{Debiased Comparative Assessment}
Let $\tilde{y}_{ij}$ represent the outcome of the comparison when considered in the opposite ordering, such that $\tilde{y}_{ij} = 1 - \hat{y}_{ji}$. For a positionally unbiased comparator, reversing the ordering should have no impact on the outcome of the comparison
\begin{equation}
    \tilde{y}_{ij} = \hat{y}_{ij} \qquad \forall \; (i, j)\in [1...N], i\ne j
\end{equation}
Systems may, however, have systematic positional biases and could for example favor the first position over the second position. To quantify the level of systematic bias, one can determine $P(A)$, the prior associated with the first position, and $P(B)$ the prior for the second position. This can be estimated for a given set of comparisons by using the statistics over all comparisons, and by calculating the fraction of times that each position is selected.
\begin{equation}
    P(A) = \frac{\sum_{i, j \in \mathcal{C}} \hat{y}_{ij}}{|\mathcal{C}|} \quad
    P(B) = \frac{\sum_{i, j \in \mathcal{C}} \tilde{y}_{ij}}{|\mathcal{C}|}
\end{equation}

\noindent When using a symmetric comparative set $\mathcal{C}$, for an unbiased system, both $P(A)$ and $P(B)$ should be 0.5 and any large deviation is symptomatic of positional bias. To address possible positional bias, one may reweight system probabilities, $\hat{p}_{ij}$, through
\begin{equation}
    \hat{p}_{ij} = \frac{\alpha \cdot p_{ij}}{\alpha \cdot p_{ij} + (1-p_{ij})}
\end{equation}
where $\alpha \in \mathbb{R}^+$ is a weight that can be set such that $P(A) = P(B) = 0.5$. Reweighting in this fashion is equivalent to, 
\begin{align}
    \hat{y}_{ij} = \begin{cases}
	1, & \text{if $p_{ij} > \tau$}\\
        0, & \text{otherwise}
    \end{cases}
\end{align}
where $\tau \in [0,1]$ is a decision threshold corresponding to $\alpha$, set such that $P(A) = P(B) = 0.5$.

\section{Experimental Setup}
\subsection{Datasets}
To investigate the general applicability of comparative assessment, we cover a range of standard NLG evaluation tasks and datasets as follows:
\vspace{1.5mm}

\noindent\textbf{SummEval} \cite{fabbri2021summeval} is a summary evaluation benchmark of 100 passages, each with 16 machine-generated summaries. Each summary is evaluated for coherency (\texttt{COH}), consistency (\texttt{CON}), fluency (\texttt{FLU}), and relevancy (\texttt{REL}).

\vspace{1.5mm}

\noindent\textbf{Podcast} \cite{manakul2022podcast} is for benchmarking podcast summary assessment methods. It contains 179 podcasts each with 15 abstractive summaries. Each summary was evaluated for its overall quality on a 4-point scale.

\vspace{1.5mm}

\noindent\textbf{TopicalChat} with the USR annotations \cite{mehri-eskenazi-2020-usr} is for benchmarking dialogue evaluation. It includes 60 dialogue contexts and six system responses per context. These responses were assessed on coherency (\texttt{COH}), continuity (\texttt{CNT}), engagingness (\texttt{ENG}), and naturalness (\texttt{NAT}).

\vspace{1.5mm}

\noindent\textbf{WebNLG} \cite{gardent-etal-2017-creating} is for benchmarking data-to-text evaluation methods. It contains 223 semantic triple groups, each paired with outputs from 8 triple-to-text generation systems. These texts were evaluated for fluency (\texttt{FLU}), grammar (\texttt{GRA}) and semantic equivalence (\texttt{SEM}).


\subsection{Base Large Language Models (LLMs)}
We investigate two families of open-source instruction-tuned LLMs. The first system is FlanT5 \cite{chung2022scaling}, T5 \cite{raffel2020exploring} that have been instruction tuned on a diverse set of 1000 NLP tasks \cite{wang2022super}. The second system is Llama2-chat \cite{touvron2023llama}, which is Llama2 tuned on instruction datasets. We investigate a range of model sizes; 220M, 770M, 3B and 11B for FlanT5, and 3B and 13B for Llama2. 


\subsection{Baselines}
\label{section:baseline}
The NLG evaluation methods can be categorized into \textit{reference-based} and \textit{reference-free}. Reference-based methods compare the output against the reference such as n-gram metrics (e.g., BLEU \cite{papineni-etal-2002-bleu} and ROUGE \cite{lin-2004-rouge}), or embedding based metrics (e.g., BERTScore \cite{zhang2019bertscore}). In contrast, reference-free methods compare the generated texts against the original source (or context for generation) directly. 

 %

\subsubsection{Bespoke Methods}
Bespoke methods require a specific data which could be supervised labels (e.g., human judgements for the summaries) or data for model training (e.g., question-answering). Although bespoke methods could work in a similar domain (e.g., developed for summarization, but applied on dialogue generation), they are not as general as zero-shot methods.

\vspace{1.5mm}
\noindent \textbf{UniEval} \cite{zhong-etal-2022-towards} convert NLG evaluation into Boolean QA problem. This method uses pre-defined schemes for selected aspects (e.g., coherence) and generates synthetic data to fine-tune a T5 system for NLG assessment. References are used for particular aspects (e.g. relevancy), and schemes/systems are bespoke for a particular attribute (though a sequentially trained system that scores multiple attributes is also explored).

\vspace{1.5mm}

\noindent \textbf{QuestEval} \cite{scialom-etal-2021-questeval} and \textbf{MQAG} \cite{manakul2023mqag} are QA-based approaches for assessing \textit{consistency} in summarization tasks. QuestEval uses extracted answer spans while MQAG represents information using multiple-choice questions. Both methods are reference-free.

\vspace{1.5mm}

\noindent \textbf{Longformer-SFT}: For podcast summarization, we follow \citet{manakul2022podcast} in using a Supervised Fine-Tuned longformer \cite{Beltagy2020Longformer} as a baseline. The input is the document and the summary, and human judgement is used as the supervised target label at training, and the performance is reported using 5-fold cross-validation. 

\subsubsection{Zero-shot Methods}
Zero-shot methods can be applied generally to any task without further training or fine-tuning. Comparative assessment is a zero-shot method.

\vspace{1.5mm}
\noindent \textbf{GPTScore} \cite{fu2023gptscore} evaluates texts using conditional language model scores. By conditioning the language model on instruction and context, GPTScore assumes that it will assign a higher probability to a high-quality generated text. 

\vspace{1.5mm}

\begin{figure}[H]
    \centering
    \includegraphics[width=0.7\columnwidth]{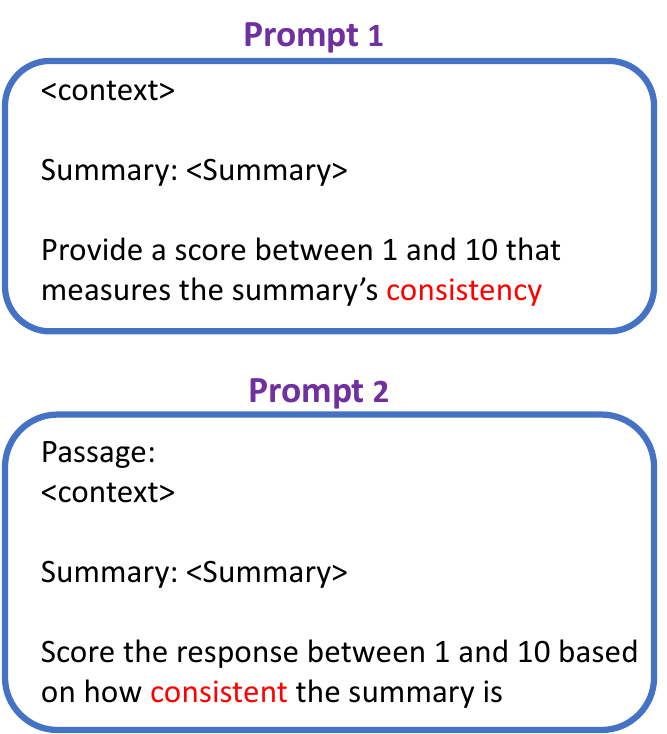}
    \caption{Scoring template 1 and template 2. Only the attribute is changed (e.g., consistent $\rightarrow$ engaging) and response description (`summary'$\rightarrow$ `response') for different tasks.}
    \label{fig:scoring_prompts}
\end{figure}

\noindent \textbf{Prompt Scoring}.
\label{section:prompt_scoring}
Another baseline is prompt-scoring. With this approach, for a particular attribute, the LLMs is asked to assess the response quality between 1-10. Simple prompts are used with the general templates shown in Figure~\ref{fig:scoring_prompts}. Prompt-scoring is run for all open-source LLMs considered (FlanT5 and Llama2), and is used as the main baseline to compare comparative assessment against. During generation, the maximum generation length is set to 5 and the temperature is set to 1.0. Similarly, ChatGPT prompt-scoring has recently been proposed in \citet{wang2023chatgpt, kocmi2023large}, which we also include as a baseline where applicable. 

\noindent \textbf{G-Eval} \cite{liu-etal-2023-g}
\label{section:prompt_scoring_2}
As an extension to prompt-scoring, G-Eval  extends standard prompt scoring by using detailed prompts and then generating a continuous score by calculating the expected score over a score range (e.g. 1-5 normalized by their probabilities). We apply G-Eval to the various base LLMs and contrast performance to the other approaches for SummEval, since the prompts for different attributes have been made publically available.\footnote{\url{https://github.com/nlpyang/geval}}







\subsection{Methodology}
Each LLM is used for both prompt-scoring and comparative assessment. For the main comparative assessment results, we consider the full set of possible comparisons, where all pairs of candidates in both permutations are compared by the framework. Comparisons are made using the prompt-based classifier (as described in \S \ref{section:comparative_decision}) using the prompt templates shown in Fig.~\ref{fig:comparative_prompts}, where the system outputs a probability for Response A and Response B. The winner of the comparison is the response with the highest probability, where candidates are then ranked in order of the win-ratio (as described in \S \ref{sec:comp_select}). For Llama2, comparative prompts are appended with `\texttt{Answer:}' while scoring prompts end with `\texttt{Score:}'. The spearman correlation between predicted scores and human judgements is used as the performance metric.

\section{Experiments}
\subsection{NLG Evaluation Results}
\textbf{Summary Assessment}: Table \ref{tab:summeval_table} analyzes the effectiveness of comparative assessment on \textbf{SummEval}, where the following observations can be made: \vspace{2mm}  

\noindent \textbf{(1)} Moderate-sized LLMs are ineffective in the prompt-scoring set-up, with the best system (FlanT5-3B) achieving Spearman correlations of 10-20. The performance difference with ChatGPT prompt-scoring implies that scoring is likely an emergent ability only effective for larger LLMs. \vspace{2mm}

\noindent \textbf{(2)} G-Eval, which uses task specific detailed prompts and continuous scores, yields significant improvements over prompt-scoring. Nonetheless, comparative assessment remains more effective than G-Eval in the majority of settings.  \vspace{2mm}

\noindent \textbf{(3)} LLMs are able to achieve considerably higher correlations in the comparative assessment set-up, with performance higher for nearly all systems. Further, comparative assessment leads to more robust performance, with most 3B+ models achieving correlations within the range of 30-50. \vspace{2mm}

\noindent \textbf{(4)} Comparative assessment enables LLMs of under 1B to perform well, with FlanT5-770M achieving moderate correlations. However, performance improves significantly when using 3B+ LLMs, although for SummEval there are diminishing (if any) performance gains by scaling up. \vspace{2mm}

\noindent \textbf{(5)} The best comparative assessment LLM (FlanT5-3B) is competitive with all other zero-shot methods, including ChatGPT scoring (an LLM with two orders of magnitude more parameters), and achieves the best correlation in 3 of the 4 aspects. 

\noindent\textbf{(6)} Comparative assessment achieves competitive performance with UniEval. Although UniEval has better overall performance, UniEval was designed for bespoke tasks and aspects (it is fine-tuned on synthetic data created for particular attributes) where the results in Tables \ref{tab:podcast_table} and \ref{tab:webnlg_table} show that UniEval has noticeable degradation in out-of-domain settings. In contrast, comparative assessment is zero-shot and general. 

%

\begin{table}[!t]
    \centering
    \small
    \begin{tabular}{l|rrrr}
        \toprule
        Approach        & COH  & CON  & FLU  & REL \\
        \midrule
        \rowcolor{Gray}
        \multicolumn{5}{l}{Baselines (\S \ref{section:baseline})} \\          
        BERTScore (w/ Ref)   & 25.9 & 19.7 & 23.7 & 34.7 \\
        QuestEval            & 18.2 & 30.6 & 22.8 & 26.8 \\
        MQAG                 & 17.0 & 28.8 & 19.3 & 16.6  \\ 
        UniEval (single-best)& 54.6 & 47.2 & 43.3 & 46.3 \\
        UniEval (continual)  & 57.5 & 44.6 & 44.9 & 42.6 \\
        GPTScore FlanT5-3B   & 47.0 & 43.6 & 42.1 & 34.4 \\
        GPTScore FlanT5-11B  & 45.6 & 43.8 & 42.4 & 34.3 \\
        GPTScore GPT3        & 40.1 & 47.5 & 41.0 & 34.3 \\
        ChatGPT scoring$^\dagger$  & 45.1 & 43.2 & 38.0 & 43.9 \\
        \rowcolor{Gray}
        \multicolumn{5}{l}{Prompt Scoring (\S \ref{section:prompt_scoring})} \\
        FlanT5-220M         & 4.0   & -0.2 & 0.2  & 2.8  \\
        FlanT5-770M         & -3.6  & -1.6 & -1.5 & -0.0 \\
        FlanT5-3B           & \bf{14.5} & \bf{19.8} & \bf{3.9} & \bf{15.2} \\
        FlanT5-11B          & 0.7   & 11.2 & 3.2  & 5.7  \\
        Llama2-chat-7B      & 8.6 & 9.0 & 1.8 & 7.8  \\
        Llama2-chat-13B     & 9.9   & 6.9  & 1.2  & 9.2  \\        
        \rowcolor{Gray}
        \multicolumn{5}{l}{G-Eval (\S \ref{section:prompt_scoring})} \\
        FlanT5-220M         & 3.6   & 0.6  & 2.7  & 8.0 \\
        FlanT5-770M         & 8.5   & 7.0  & 15.3 & 24.1  \\
        FlanT5-3B           & 10.5	& 29.1 & 9.8  & 23.8 \\
        FlanT5-11B          & 19.2	& 29.3 & 20.7 & 35.8 \\
        Llama2-chat-7B      & 28.2	& 29.4 & \bf{23.0} & 27.4 \\
        Llama2-chat-13B     & \bf{53.2}  & \bf{33.7} & 16.5 & \bf{38.3} \\
        \rowcolor{Gray}
        \multicolumn{5}{l}{Comparative Assessment (\S \ref{section:comparative_assessment})} \\        
        FlanT5-220M         & 4.0  & -0.2 & 0.2  & 2.8   \\
        FlanT5-770M         & 29.8 & 26.3 & 20.6 & 35.1  \\
        FlanT5-3B           & \bf{51.2} & \bf{47.1} & \bf{32.5} & 44.8  \\
        FlanT5-11B          & 44.2 & 37.2 & 30.2 & 43.4  \\
        Llama2-chat-7B     & 27.9 & 24.6 & 20.2 & 35.6  \\
        Llama2-chat-13B    & 40.9 & 39.9 & 30.8 & \bf{45.3}  \\
        \bottomrule
    \end{tabular}
    \caption{Spearman correlation coefficient for \textbf{SummEval}, averaged over both prompts per system (for prompt-scoring and comparative). $^\dagger$ChatGPT performance is quoted from \citet{wang2023chatgpt}, which use more detailed scoring prompts.} 
    \label{tab:summeval_table}
\end{table}

\begin{table}[!t]
    \centering
    \small
    \begin{tabular}{l|cc}
        \toprule
        Approach  & System-lvl & Summary-lvl \\
        \midrule
        \rowcolor{Gray}
        \multicolumn{3}{l}{Baselines (\S \ref{section:baseline})} \\          
        BERTScore (w/ Ref) &73.9 &25.1\\ 
        UniEval (continual) & 42.0 & 22.8 \\
        QuestEval &42.5 &20.4\\
        MQAG      &77.9 &12.6\\
        Longformer-SFT &89.6 &19.6 \\
        \rowcolor{Gray}
        \multicolumn{3}{l}{Prompt Scoring (\S \ref{section:prompt_scoring})} \\
        Llama2-chat-7B          &88.5 & 2.6 \\
        Llama2-chat-13B         &80.0 & 25.3 \\ 
        \rowcolor{Gray}
        \multicolumn{3}{l}{Comparative Assessment (\S \ref{section:comparative_assessment})} \\        
        Llama2-chat-7B          &88.2 & 37.4 \\
        Llama2-chat-13B         &97.1 & 45.5\\
        \bottomrule
    \end{tabular}
    \caption{Spearman correlation coefficient for \textbf{Podcast}.} 
    \label{tab:podcast_table}
\end{table}

\vspace{1.5mm}
\noindent\textbf{Podcast Assessment}: When considering podcast summarization with long inputs of over 5k tokens on average, only Llama2 models (which have a limit of 4k tokens) were used (as FlanT5 has a limit of 1k tokens). Table \ref{tab:podcast_table} shows that comparative assessment yields highly impressive performance for long-spoken summarization, with comparative assessment out-competing all other baselines. Further, although prompt-scoring has good system-level correlations, the lack of granularity leads to poor summary-level performance. 



\begin{table}[!t]
    \centering
    \small
    \begin{tabular}{l|rrrr}
        \toprule
        Approach   & COH  & CNT  & ENG  & NAT \\
        \midrule
        \rowcolor{Gray}
        \multicolumn{5}{l}{Baselines (\S \ref{section:baseline})} \\          
        UniEval (single-best)& 60.7 & -    & 59.6 & 54.7 \\
        UniEval (continual)  & 61.3 & -    & 60.5 & 44.4 \\
        GPTScore GPT3        & 56.9 & 32.9 & 49.6 & 52.4 \\ 
        ChatGPT scoring$^\dagger$ & 54.7 & 57.7 & 37.9  & 58.0 \\
        \rowcolor{Gray}
        \multicolumn{5}{l}{Prompt Scoring (\S \ref{section:prompt_scoring})} \\
        FlanT5-220M         & -2.2 & 0.2  & -8.4 & 2.1  \\
        FlanT5-770M         & 3.7  & 3.1  & -4.3 & 3.8  \\
        FlanT5-3B           & \bf{31.9} & \bf{28.8} & 17.4 & \bf{23.7} \\
        FlanT5-11B          & 15.3 & 8.0  & 4.3  & 24.3 \\
        Llama2-chat-7B     & 16.4 & 17.0 & 20.6 & 21.4 \\
        Llama2-chat-13B    & 21.7 & 19.9 & \bf{31.4} & 23.2  \\
        \rowcolor{Gray}
        \multicolumn{5}{l}{Comparative Assessment (\S \ref{section:comparative_assessment})} \\        
        FlanT5-220M         & -0.3 & 8.2  & -10.5 & 2.2  \\
        FlanT5-770M         & 38.5 & 36.3 & 25.3  & 35.3 \\
        FlanT5-3B           & 49.4 & \bf{49.4} & 37.3  & 47.4 \\
        FlanT5-11B          & \bf{54.3} & 42.2 & 54.7  & \bf{54.2} \\
        Llama2-chat-7B     & 28.9 & 33.7 & 36.1 & 30.3  \\
        Llama2-chat-13B    & 32.4 & 43.2 & \bf{55.5} & 33.5  \\
        \bottomrule
    \end{tabular}
    \caption{Spearman correlation coefficient for \textbf{TopicalChat}. $^\dagger$ChatGPT is prompted using our prompt-scoring prompts.} 
    \label{tab:topicalchat_table}
\end{table}

\vspace{1.5mm}
\noindent{\textbf{Dialogue Assessment}}: Next, we analyze comparative assessment on TopicalChat, for evaluating conversational responses. Table \ref{tab:topicalchat_table} shows similar findings for TopicalChat as to those in SummEval, where comparative assessment again outperforms the correlations seen from prompt-scoring. 

\vspace{1.5mm}
\noindent\textbf{Data-to-Text Assessment}: For data-to-text generation, the context is highly abstract and is a list of triples in the form of (object, relation, subject). This makes assessing the semantics challenging, as the LLM needs to parse and understand semantic triples. Table \ref{tab:webnlg_table} shows that understanding triples is an emergent ability of LLMs, where for grammar and fluency the correlations are quite similar between the 3B and 11B/13B systems, however for semantic understanding, the 10B+ systems highly outcompete the 3B+ systems. Note that when evaluating UniEval, we used the closest attribute that they designed for, which was naturalness for both.

\subsection{Positional Bias}
\label{sec:pos_bias}
We investigate whether the comparative prompts have any implicit positional bias, and whether systems prefer the first/second position. Table \ref{tab:summeval_bias} shows the fraction of comparisons that selected the candidate in the first position for SummEval. Since all comparisons in both permutations are considered, this fraction should be 0.50 for an unbiased system. However, we observe considerably high bias, with some set-ups even selecting the first option 80\% of the time. Further, we observe that larger systems appear to be more susceptible to bias than smaller systems, which may explain the similarity in performance for the 3B and 11B/13B systems in the previous main results. Similar results for other datasets are provided in Appendix~\ref{sec:app_bias}

\begin{table}[!t]
    \centering
    \small
    \begin{tabular}{l|rrr}
        \toprule
        Approach   & FLU & GRA & SEM \\
        \midrule
        \rowcolor{Gray}
        \multicolumn{4}{l}{Baselines (\S \ref{section:baseline})} \\          
        BLEU           & 36.3 & 34.7 & 50.3\\
        METEOR         & 44.3 & 42.9 & 62.7 \\
        NLI Model$^*$  & -  & -  & 63.7 \\
        UniEval (continual) & 21.7 & 16.3 & - \\
        \rowcolor{Gray}
        \multicolumn{4}{l}{Prompt Scoring (\S \ref{section:prompt_scoring})}  \\
        FlanT5-220M        & 18.5 & 17.4 & 8.0  \\
        FlanT5-770M        & 14.5 & 13.6 & 17.1 \\
        FlanT5-3B          & \bf{30.8} & \bf{32.7} & \bf{38.5} \\
        FlanT5-11B         & -0.7 & 6.9  & 20.8 \\
        Llama2-chat-7B    & 3.8  & 2.4  & 17.0 \\
        Llama2-chat-13B   & 1.8  & 0.5  & 5.6 \\
        \rowcolor{Gray}
        \multicolumn{4}{l}{Comparative Assessment (\S \ref{section:comparative_assessment})} \\        
        FlanT5-220M        & -13.6 & -17.9 & 0.1 \\
        FlanT5-770M        & 36.2 & 35.2   & 11.4 \\
        FlanT5-3B          & 40.6 & 41.4   & 12.8 \\
        FlanT5-11B         & 41.4 & 44.8 & 52.4 \\
        Llama2-chat-7B    & 22.9 & 37.8 & -5.3\\
        Llama2-chat-13B   & \bf{44.9} & \bf{45.1} & \bf{53.5} \\
        \bottomrule
    \end{tabular}
    \caption{Spearman correlation coefficient for \textbf{WebNLG}. $^*$Quoted from the NLI method with the backoff template in \citet{dusek-kasner-2020-evaluating}.} 
    \label{tab:webnlg_table}
\end{table}

\begin{table}[!h]
    \small
    \centering
    \begin{tabular}{cc|cccc}
        \toprule
        System       &Prompt & COH & CON & FLU & REL \\
        \midrule
        FlanT5      & 1 & 0.37 & 0.46 & 0.39 & 0.41 \\
        3B          & 2 & 0.43 & 0.47 & 0.40 & 0.44 \\
        \midrule
        FlanT5      & 1 & 0.18 & 0.20 & 0.13 & 0.23 \\ 
        7B          & 2 & 0.24 & 0.24 & 0.17 & 0.26 \\ 
        \midrule
        Llama2-chat & 1 & 0.41 & 0.17 & 0.26 & 0.18 \\
        7B          & 2 & 0.68 & 0.56 & 0.48 & 0.45 \\
        \midrule 
        Llama2-chat & 1 & 0.31 & 0.37 & 0.18 & 0.32 \\
        13B         & 2 & 0.29 & 0.30 & 0.19 & 0.26 \\
       \bottomrule
    \end{tabular}
    \caption{Positional bias $P(A)$ for both prompt templates, for various systems in the comparative setup on SummEval.}
    \label{tab:summeval_bias}
\end{table}

\begin{table*}[t]
    \centering
    \small
    \begin{tabular}{lc|rrrr|rrrr|rr|r}
        \toprule
        \multirow{2}{*}{System} &\multirow{2}{*}{Debias} & \multicolumn{4}{c|}{SummEval} & \multicolumn{4}{c|}{TopicalChat} & \multicolumn{2}{c|}{WebNLG} &\multirow{2}{*}{Avg.}  \\
                      & &COH  & CON & FLU & REL & COH  & CNT  & ENG  & NAT & FLU & GRA  & \\
        \midrule
        \multirow{2}{*}{FlanT5-3B}           & \xmark & 51.2 & 47.1 & 32.5 & 44.8 & 49.4 & 49.4 & 37.3 & 47.4 & 41.0 & 41.8 & 44.2 \\
                            & \cmark & 51.8 & 46.9 & 33.0 & 45.3 & 49.6 & 50.2 & 38.0 & 46.3 & 40.7 & 42.3 & 44.4 \\
        \midrule
        \multirow{2}{*}{FlanT5-11B}          & \xmark & 44.2 & 37.2 & 30.2 & 43.4 & 54.3 & 42.2 & 54.7 & 54.2 & 41.4 & 44.8 & 44.7 \\
                            & \cmark & 45.3 & 39.7 & 30.7 & 44.7 & 57.2 & 59.5 & 59.5 & 58.8 & 44.5 & 44.6 & 48.5 \\
        \midrule
        \multirow{2}{*}{Llama2-chat-7B}     & \xmark & 29.4 & 24.6 & 19.7 & 35.2 & 28.2 & 33.1 & 36.3 & 28.7 & 22.9 & 37.8 & 29.6 \\
                            & \cmark & 28.8 & 24.8 & 19.7 & 35.5 & 29.1 & 34.5 & 39.7 & 28.5 & 24.3 & 37.1 & 30.2 \\
        \midrule
        \multirow{2}{*}{Llama2-chat-13B}    & \xmark & 40.9 & 39.9 & 30.8 & 45.3 & 32.4 & 43.2 & 55.5 & 33.5 & 44.9 & 45.1 & 41.2 \\
                            & \cmark & 42.8 & 40.3 & 31.9 & 47.1 & 32.5 & 44.5 & 56.9 & 38.4 & 45.9 & 43.7 & 42.4 \\
        \bottomrule
    \end{tabular}
    \caption{Spearman correlation coefficient on different aspects of the NLG evaluation tasks, averaged over all prompts considered, using all pairs and ordering considered (i.e. full matrix comparisons).} 
    \label{tab:debias_table}
\end{table*} 

\subsection{Debiasing}
The previous section demonstrates that comparative assessment exhibits positional bias which may impact system decisions. We therefore investigate whether debiasing can improve evaluation performance. Table \ref{tab:debias_table} shows standard and debiased LLM comparative assessment performance for the considered tasks and scores, with WebNLG \texttt{SEM} and Podcast omitted due to the required emergent ability and large context length respectively. We observe that debiasing can lead to performance boosts, where we note that the prompts which have a high bias (seen in Table \ref{tab:summeval_bias} and Table \ref{tab:app_bias_table} in the appendix) benefit most from debiasing. In particular, for TopicalChat we observe large gains for the FlanT5-11B system, which enables state-of-the-art performance. To explain why debiasing can lead to large performance boosts, consider a very biased system where the first response is always selected as better. Although over both permutations the system is unbiased for any comparison, the bias in the system will cause the system to assume that all candidates are of the same quality. By reducing the bias of each comparison, the system may be able to pick up subtler quality differences between the samples. 

\subsection{Comparative Accuracy}
\begin{table}[h]
    \small
    \tabcolsep=1.6mm
    \centering
    \begin{tabular}{lc|cccc}
        \toprule
        System       & Debias & COH & CON & FLU & REL \\
        \midrule
        \multirow{2}{*}{FlanT5-3B}    
                    & \xmark & 68.6 & 82.0 & 68.2 & 67.2 \\
                    & \cmark & 69.8 & 82.1 & 68.8 & 67.8 \\
        \midrule
        \multirow{2}{*}{FlanT5-11B}    
                    & \xmark & 61.6 & 70.3 & 60.3 & 63.3 \\ 
                    & \cmark & 66.2 & 76.7 & 65.9 & 67.4 \\ 
        \midrule
        \multirow{2}{*}{Llama2-chat-7B}    
                    & \xmark & 59.6 & 63.8 & 59.6 & 61.0 \\
                    & \cmark & 60.3 & 65.7 & 60.4 & 63.1 \\
        \midrule 
        \multirow{2}{*}{Llama2-chat-13B}    
                    & \xmark & 62.6 & 75.4 & 61.1 & 65.4 \\
                    & \cmark & 65.8 & 76.9 & 67.2 & 68.5 \\
       \bottomrule
    \end{tabular}
    \caption{Accuracy of the comparative systems, at a comparison level, for SummEval.}
    \label{tab:acc_debias}
\end{table}
\noindent  One can also measure the accuracy of the comparative system at a comparison level. Table \ref{tab:acc_debias} shows the pairwise comparison accuracy for Summeval, over all candidate pairs where the true score of the candidate response varies. We observe accuracies between 60-80\% across all tasks and observe that debiasing can substantially increase accuracy. This highlights that LLMs are able to compare the quality of responses fairly well, though the moderately sized LLMs may not always select the best response (with respect to labels).

\subsection{Self-Consistency}
SummEval has 16 summaries per context which leads to 240 possible comparisons. If one were to instead randomly sample $N$ outputs and consider all $N\!\cdot\!(N\!-\!1)$ comparisons, how consistent would the rankings with the subset of systems be with respect to the final predicted rankings? Table~\ref{tab:self consistency} illustrates the self-consistency measured by the accuracy when comparing pairs, and demonstrates that even when using few outputs, the model is very consistent to the final rankings that would be achieved by using many more examples.

\begin{table}[H]
    \centering
    \small
    \renewcommand\tabcolsep{5pt}
    \begin{tabular}{c|ccccccc}
        \toprule
              & 2	& 3 & 4 & 6 & 8 & 12& 16 \\
        \midrule
        Final & 84.0 & 88.3 & 90.7 & 93.7 & 95.5 & 98.0 & 100\\
        Gold  & 68.0 & 69.1 & 69.7 & 70.3 & 70.6 & 70.8 & 70.9 \\
        \bottomrule
    \end{tabular}
    \caption{Accuracy when using fewer systems with respect to final rankings (using all 16 systems) and the ground truth labels. Results shown for Summeval \texttt{COH} using FlanT5-xl.}
    \label{tab:self consistency}
\end{table}

\subsection{Subset of Comparisons}
\begin{figure}[h]
    \centering    \includegraphics[width=\linewidth,keepaspectratio]{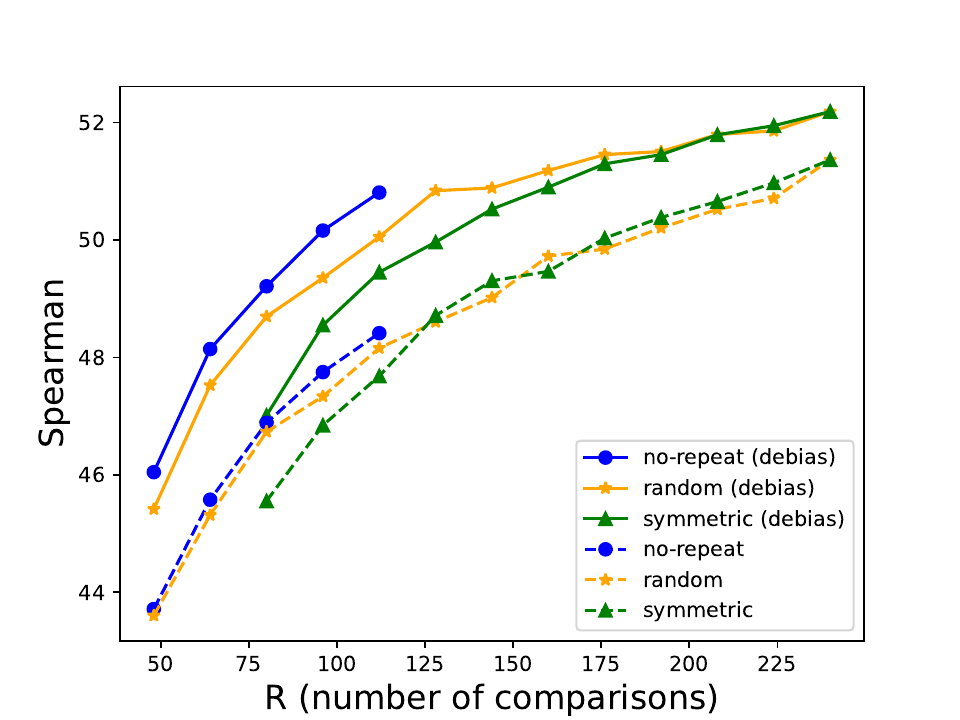}
    \caption{FlanT5-3B performance for SummEval \texttt{COH} when a subset of the comparisons are selected by either random, no-repeat or symmetric (as described in \S \ref{sec:comp_select}). For no-repeat, each pair is compared once, hence has a smaller maximum $R$.} 
    \label{fig:efficient}
\end{figure}

Due to $O(N^2)$ number of comparisons required for the full comparison matrix, it might be practical to only consider a subset of comparisons. Figure \ref{fig:efficient} shows the downstream Spearman correlation for SummEval coherency, when averaged over 50 runs, for different comparison selection strategies. Of the three schemes, we observe that for small $R$ (i.e. less than half the total number of comparisons) selecting comparisons with no repeats leads to a marginal improvement over random selection. Further, by using the symmetric selection scheme, despite the number of comparisons being half that of no-repeat (although each comparison is done twice, once in each permutation), interestingly there is only a performance difference of 1 in terms of Spearman. Finally, we observe that debiasing can be very effective in efficient set-ups, and leads to larger benefits when the number of comparisons is small. Equivalent plots for other tasks/scores can be found in Appendix \ref{sec:app_efficient}. 

\section{Conclusions}
This paper investigates LLM comparative assessment, a simple zero-shot approach to NLG evaluation. We demonstrate that for moderately sized LLMs, comparative assessment outperforms absolute scoring, and is an effective automatic assessment, achieving near state-of-the-art performance for a range of NLG evaluation tasks. Furthermore, we show that LLMs are prone to have positional bias that could impact their decisions, however, we introduce a simple debiasing approach that leads to performance boosts, especially for biased systems.

\section*{Limitations}
\noindent  \textit{Computational Cost}. The comparative assessment framework with the full set of comparisons uses $N \cdot (N-1)$ comparisons, which for large $N$ can be computationally prohibitive. This paper investigated datasets with at most 16 candidates, and may not scale when more candidates are required.

\vspace{1.5mm}
\noindent  \textit{Base LLMs}. The empirical findings are for LLMs of up to 13B parameters. By using larger models (with 100B+ parameters) one may expect further performance improvements. However, due to API costs and the O($N^2$) number of comparisons, results are limited to open-source LLMs.

\vspace{1.5mm}
\noindent \textit{Selection of the subset of comparisons}. For our comparison selection scheme, this work only considered static selection schemes. Future work may investigate dynamic selection schemes, either by considering sorting algorithms or ELO competition schemes, and methods similar to those studied in information retrieval by \citet{qin2023large}.

\section*{Ethics Statement}
For some tasks/datasets, comparative assessment could be ineffective and have poor generalisation over the task. Deploying machine learning classifiers in real-world classification settings has many associated risks, and careful analysis should be made before deploying such systems. Misuse/overconfidence in the approach may lead to mistrust of users towards LLM solutions.

\section*{Acknowledgements}
This paper reports on research supported by Cambridge University Press \& Assessment (CUP\&A), a department of The Chancellor, Masters, and Scholars of the University of Cambridge. This research is further supported by the Cambridge International \& St John’s College scholarship.

\bibliography{anthology,custom}

\appendix

\newpage
\onecolumn
\section{Additional Results}
\subsection{Partial Comparison Curves}
\label{sec:app_efficient}

\begin{figure}[H]
    \centering
    \begin{subfigure}[t]{0.33\textwidth}
        \centering
        \includegraphics[width=\linewidth,keepaspectratio]{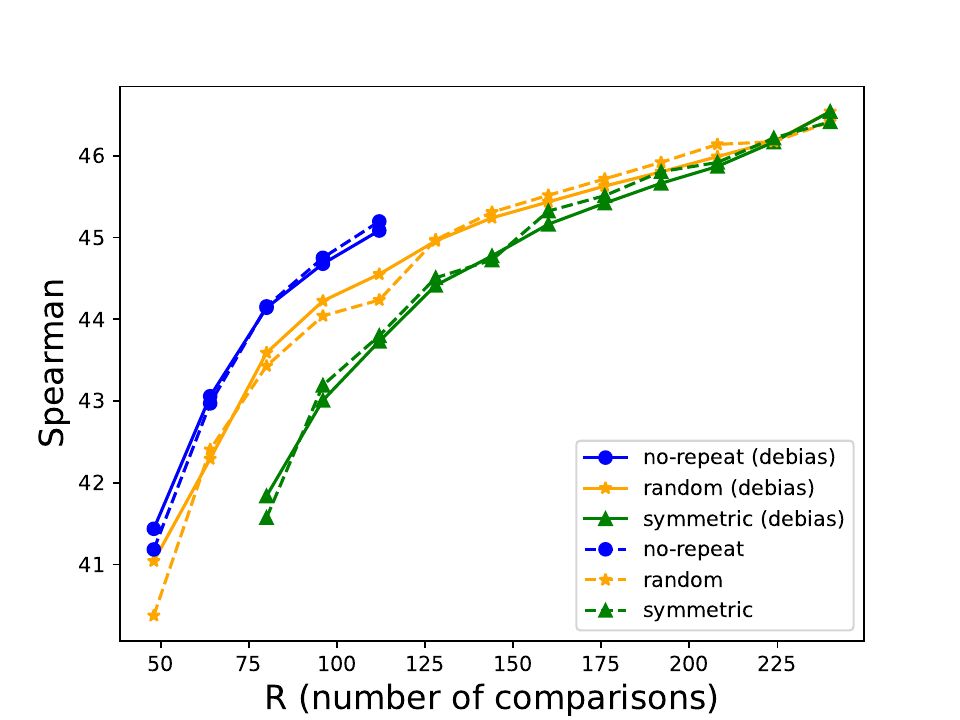}
        \caption{FlanT5-3B, SummEval, \texttt{CON}}
    \end{subfigure}%
    ~ 
    \begin{subfigure}[t]{0.33\textwidth}
        \centering
        \includegraphics[width=\linewidth,keepaspectratio]{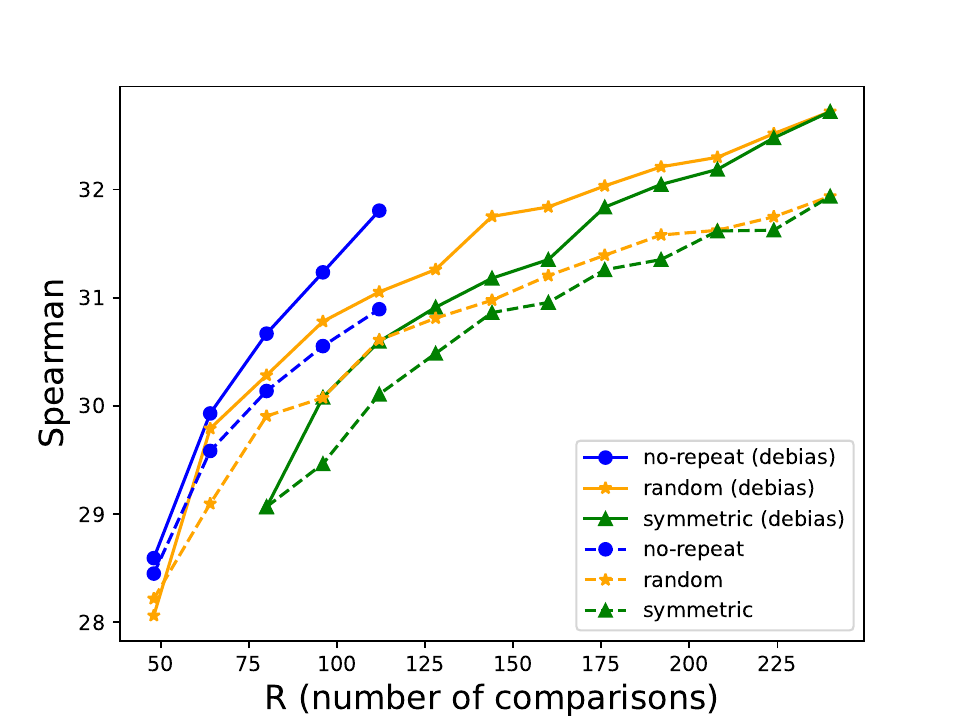}
        \caption{FlanT5-3B, SummEval, \texttt{FLU}}
    \end{subfigure}%
    \begin{subfigure}[t]{0.33\textwidth}
        \centering
        \includegraphics[width=\linewidth,keepaspectratio]{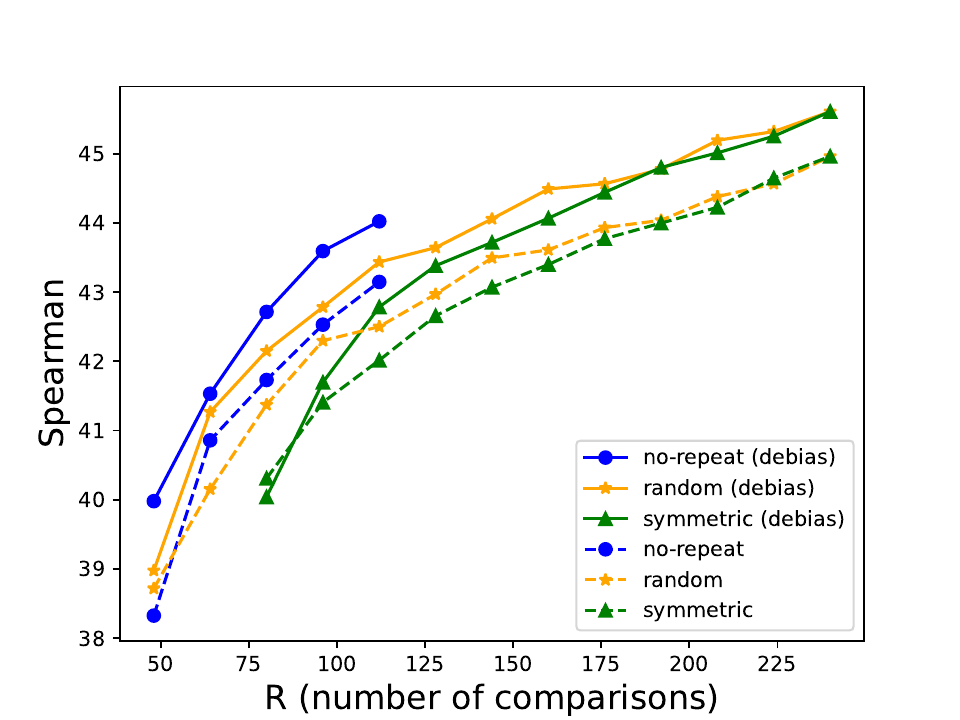}
        \caption{FlanT5-3B, SummEval, \texttt{REL}}
    \end{subfigure}
    ~
    \begin{subfigure}[t]{0.33\textwidth}
        \centering        \includegraphics[width=\linewidth,keepaspectratio]{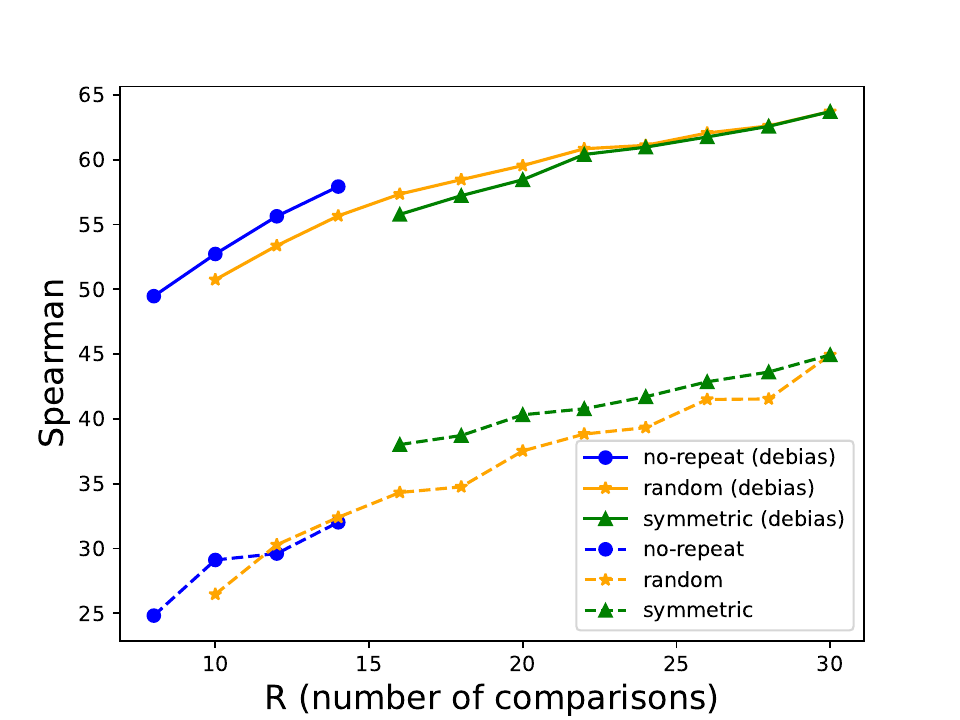}
        \caption{FlanT5-11B, TopicalChat, \texttt{COH}}
    \end{subfigure}%
    ~
    \begin{subfigure}[t]{0.33\textwidth}
        \centering
        \includegraphics[width=\linewidth,keepaspectratio]{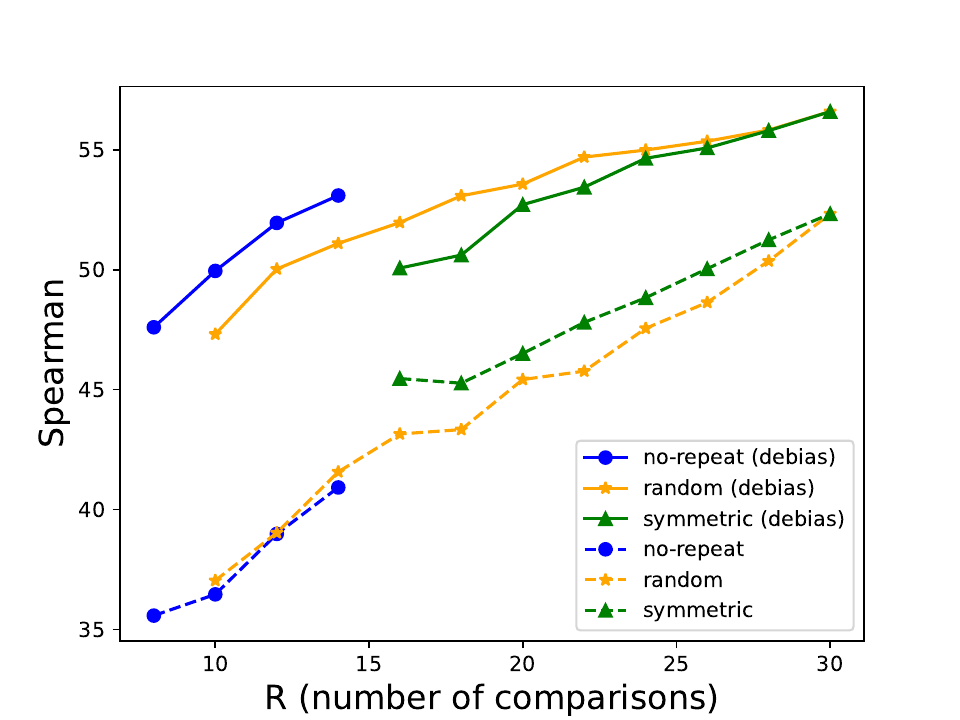}
        \caption{FlanT5-11B, TopicalChat, \texttt{ENG}}
    \end{subfigure}%
    ~
    \begin{subfigure}[t]{0.33\textwidth}
        \centering
        \includegraphics[width=\linewidth,keepaspectratio]{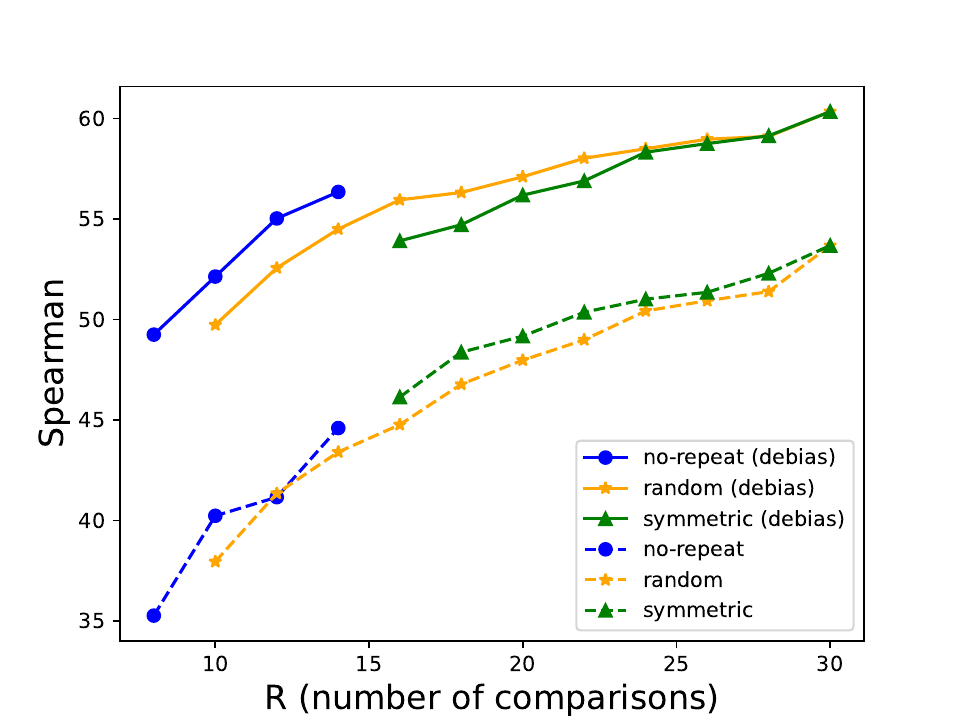}
        \caption{FlanT5-11B, TopicalChat, \texttt{NAT}}
    \end{subfigure}
        ~
    \begin{subfigure}[t]{0.33\textwidth}
        \centering
        \includegraphics[width=\linewidth,keepaspectratio]{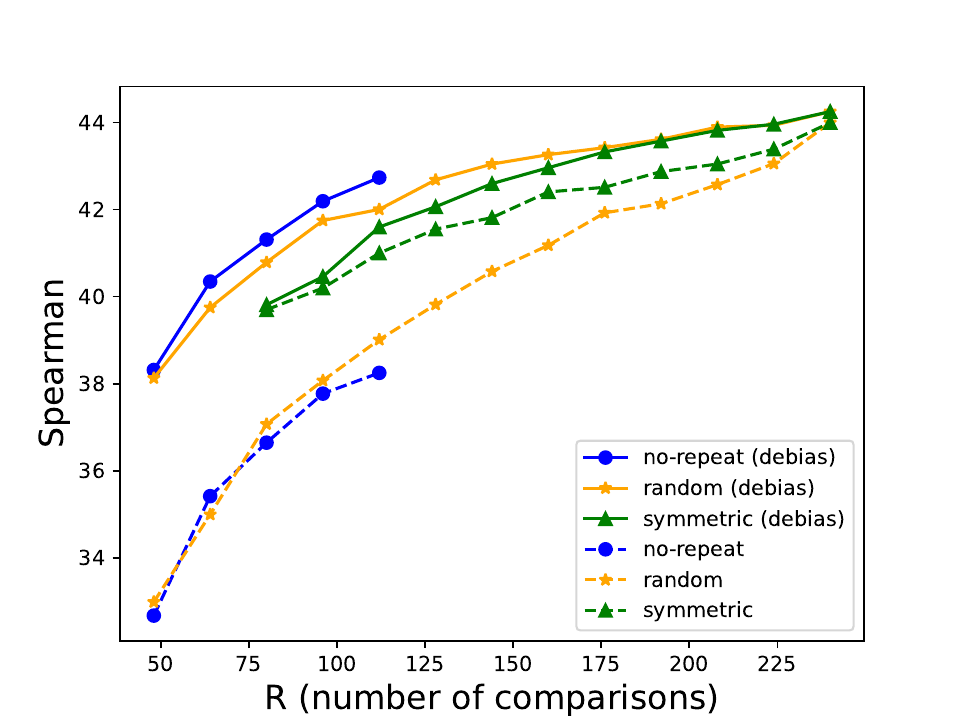}
        \caption{FlanT5-11B, SummEval, \texttt{REL}}
    \end{subfigure}%
    ~
    \begin{subfigure}[t]{0.33\textwidth}
        \centering
        \includegraphics[width=\linewidth,keepaspectratio]{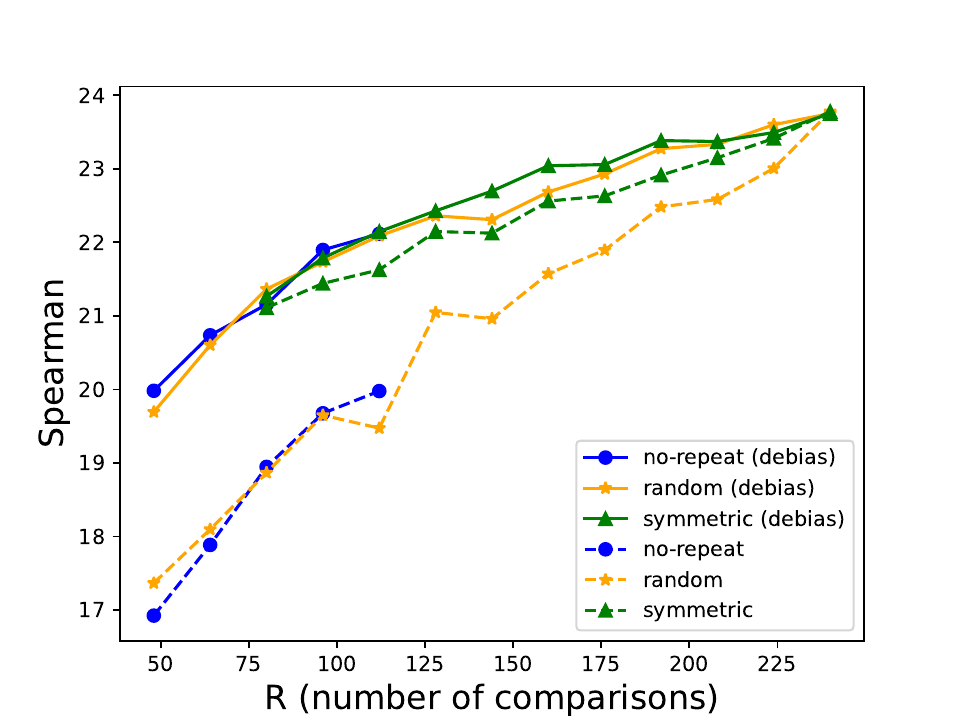}
        \caption{Llama-chat-7B, SummEval, \texttt{CON}}
    \end{subfigure}%
    ~
    \begin{subfigure}[t]{0.33\textwidth}
        \centering
        \includegraphics[width=\linewidth,keepaspectratio]{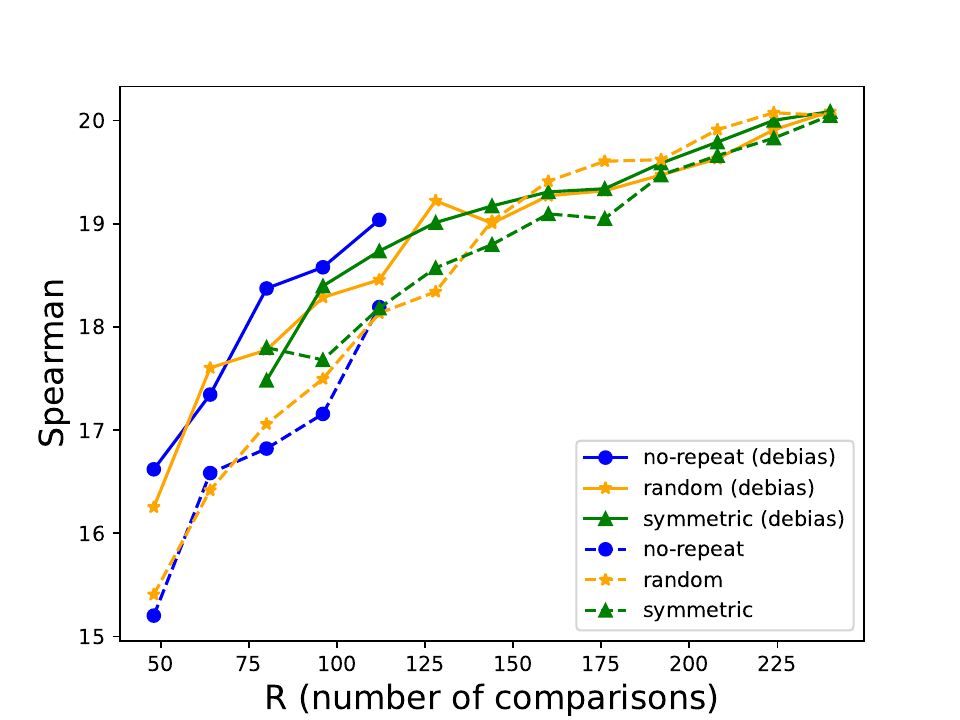}
        \caption{Llama-chat-13B, SummEval, \texttt{FLU}}
    \end{subfigure}
    \caption{Assessment Performance when only a subset of comparisons are considered (extending the results of Figure \ref{fig:efficient}). Multiple different base LLMs, datasets and scores and displayed.}
    \label{fig:efficient-app}
\end{figure}

\subsection{Positional Bias}
\label{sec:app_bias}

\begin{table*}[h!]
    \centering
    \tabcolsep=1.75mm
    \small
    \begin{tabular}{cc|rrrr|rrrr|rrr|c}
        \toprule
        \multirow{2}{*}{System} &\multirow{2}{*}{prompt} & \multicolumn{4}{c|}{SummEval} & \multicolumn{4}{c|}{TopicalChat} & \multicolumn{3}{c|}{WebNLG} &\multirow{2}{*}{Podcast}  \\
                      & &COH  & CON & FLU & REL & COH  & CNT  & ENG  & NAT & FLU & GRA & SEM \\
        \midrule
        FlanT5 & 1  & 0.37 & 0.46 & 0.41 & 0.42 & 0.47 & 0.44 & 0.50 & 0.49 & 0.46 & 0.41 & 0.89 & - \\
        3B     & 2  & 0.43 & 0.47 & 0.42 & 0.44 & 0.46 & 0.44 & 0.47 & 0.47 & 0.38 & 0.36 & 0.85 & - \\
        \midrule
        FlanT5 & 1  & 0.18 & 0.25 & 0.16 & 0.23 & 0.25 & 0.17 & 0.27 & 0.26 & 0.15 & 0.19 & 0.56 & - \\
        11B        & 2 & 0.24 & 0.29 & 0.19 & 0.26 & 0.27 & 0.13 & 0.29 & 0.31 & 0.19 & 0.21 & 0.42 & - \\
        \midrule
        Llama2-chat & 1  & 0.41 & 0.21 & 0.28 & 0.18 & 0.57 & 0.26 & 0.25 & 0.36 & 0.36 & 0.53 & 0.98 & 0.33 \\
        7B          & 2  & 0.68 & 0.57 & 0.50 & 0.45 & 0.56 & 0.37 & 0.22 & 0.35 & 0.37 & 0.48 & 0.90 & 0.24 \\
        \midrule
        Llama2-chat & 1  & 0.31 & 0.43 & 0.20 & 0.32 & 0.69 & 0.73 & 0.67 & 0.74 & 0.23 & 0.38 & 0.50 & 0.22 \\
        13B & 2  & 0.29 & 0.37 & 0.22 & 0.26 & 0.65 & 0.65 & 0.62 & 0.68 & 0.28 & 0.40 & 0.29 & 0.40 \\
        \bottomrule
    \end{tabular}
    \caption{Fraction of comparisons where the candidate in the first position was selected by the LLM when using the full (symmetric) set of comparisons. The bias is presented for both prompts, over all datasets and scores, extending the results in Table \ref{tab:summeval_bias}.} 
    \label{tab:app_bias_table}
\end{table*} 

\subsection{Accuracy of Pairwise Comparisons}
\begin{table}[h]
    \centering
    \tabcolsep=1.75mm
    \small
    \begin{tabular}{cc|rrrr|rrrr|rrr|c}
        \toprule
        \multirow{2}{*}{System} &\multirow{2}{*}{debias} & \multicolumn{4}{c|}{SummEval} & \multicolumn{4}{c|}{TopicalChat} & \multicolumn{3}{c|}{WebNLG} &\multirow{2}{*}{Podcast}  \\
                      & &COH  & CON & FLU & REL & COH  & CNT  & ENG  & NAT & FLU & GRA & SEM \\
        \midrule
        FlanT5 & \xmark  & 68.6 & 82.0 & 68.2 & 67.2 & 75.3 & 71.0 & 65.6 & 70.3 & 66.2 & 65.5 & 51.8 & - \\
        3B     &\cmark & 69.8 & 82.1 & 68.8 & 67.8 & 75.4 & 72.2 & 65.6 & 69.9 & 66.7 & 66.6 & 51.3 & - \\
        \midrule
        FlanT5 & \xmark  & 61.6 & 70.3 & 60.3 & 63.3 & 70.0 & 60.5 & 68.0 & 68.9 & 60.8 & 62.7 & 69.6 & - \\
        11B        &\cmark  & 66.2 & 76.7 & 65.9 & 67.4 & 76.6 & 74.2 & 74.4 & 74.7 & 67.6 & 67.3 & 69.9 & - \\
        \midrule
        Llama2-chat &\xmark & 59.6 & 63.8 & 59.6 & 61.0 & 64.0 & 62.0 & 61.0 & 60.4 & 56.6 & 61.1 & 48.3 & 63.4 \\
        7B          &\cmark  & 60.3 & 65.7 & 60.4 & 63.1 & 64.0 & 64.3 & 65.9 & 61.6 & 57.1 & 61.1 & 50.2 & -\\
        \midrule
        Llama2-chat &\xmark  & 62.6 & 75.4 & 61.1 & 65.4 & 64.5 & 66.8 & 72.0 & 62.3 & 64.7 & 67.6 & 67.3 & 70.3 \\
        13B &\cmark  & 65.8 & 76.9 & 67.2 & 68.5 & 65.9 & 69.4 & 73.8 & 65.2 & 66.7 & 67.4 & 68.9 & - \\
        \bottomrule
    \end{tabular}
    \caption{Accuracy of pairwise comparisons of all candidates which differ in true value. Accuracies are shown for all datasets and scores, extending the results of Table \ref{tab:debias_table}.} 
    \label{tab:app_acc_table}
\end{table} 

\begin{multicols}{2}
\section{Alternate Ranking Strategies}
\label{appendix:decoding}
In the main paper, we only consider the win ratio as an approach of converting comparisons to ranks, due to win-ratio being simple and intuitive. However alternate ranking strategies are possible; a well-motivated decoding approach is to select the ranks with the highest probability given the observed comparisons. By Bayes' theorem, this is equivalent to finding the ranks that maximise the likelihood of the observations.
\begin{align}
    \hat{r}_{1:N} &= \argmax{r_{1:N}} P(\mathcal{C}|r_{1:N})
\end{align}
For a set of ranks $r_{1:N}$, let $z_{ij}\!=\!\mathbbm{1}(r_i\!\!<\!\!r_j) \!\in\! \{0, 1\}$, i.e. whether the ranks imply $x_i$ is better than $x_j$. Given the probability of each comparison, the \textbf{likelihood} of the ranks can be defined as 
\begin{align}
    P(\mathcal{C}|r_{1:N}) = \prod_{(i, j) \in \mathcal{C}} \big{(} p_{ij}^{z_{ij}} + (1-p_{ij})^{1-z_{ij}} \big{)}
\end{align}
If only hard decisions are available (i.e. the probabilities are not), then one can instead approximate the likelihood and find the ranks that maximise the \textbf{approximate-likelihood}. 
\begin{align}
    P(\mathcal{C}|r_{1:N}) = \prod_{(i, j) \in \mathcal{C}} P(\hat{y}_{ij} | z_{ij})
\end{align}
Since $\hat{y}_{ij}\in\{0,1\}$ and $z_{ij}\in\{0,1\}$, there are 4 conditional probabilities $P(\hat{y}_{ij} | z_{ij})$. Setting one probability will set the other 3, which can be estimated with the system's comparative statistics. \\

\subsection{Initial Results}
Table \ref{tab:decoding} presents initial results for FlanT5-3B on Summeval, comparing the maximum likelihood ranking to the win ratio approach. The initial finding was that performance was similar between the two conversion schemes. However, it's worth noting that minimizing the objective function poses intractability challenges, necessitating an approximate greedy search. For the sake of simplicity, our main paper focused on the win-ratio method, while future research may explore more advanced conversion strategies.

\begin{table}[H]
    \centering
    \begin{tabular}{c|cccc}
        \toprule
                    & \multicolumn{4}{c}{SummEval} \\
                    & COH  & CON  & FLU  & REL  \\
                    \midrule
        win-loss     & 51.4 & 46.4 & 31.9 & 45.0 \\
        likelihood   & 51.7 & 46.0 & 31.5 & 44.7 \\
        \bottomrule
    \end{tabular}
    \caption{Spearman correlation when the comparisons are converted using either win-ratio or maximum likelihood, for FlanT5-3B on SummEval.}
    \label{tab:decoding}
\end{table}
\vspace{6cm}
\phantom{Pembroke's going up}
\end{multicols}
\end{document}